\documentclass{article}

    \usepackage[preprint]{neurips_2020}

\usepackage[]{neurips_2020}
\usepackage[utf8]{inputenc} 
\usepackage[T1]{fontenc}    
\usepackage{hyperref}       
\usepackage{url}            
\usepackage{booktabs}       
\usepackage{amsfonts}       
\usepackage{nicefrac}       
\usepackage{microtype}      

\usepackage{microtype}
\usepackage{graphicx}
\usepackage{subfigure}
\usepackage{amsfonts, amsmath}
\usepackage{amssymb}
\usepackage{comment}
\usepackage{booktabs}
\newcommand{\printfnsymbol}[1]{%
  \textsuperscript{\@fnsymbol{#1}}%
}
\newcommand{\E}{\mathbb{E}}
\title{GABO: Graph Augmentations with\\ Bi-level Optimization}

\author{%
  Heejung W. Chung \thanks{Equal contribution}\\
  Dept of Computer Science\\
  Stanford University \\
  \texttt{hchung98@stanford.edu} \\
  \And
  Avoy Datta \footnotemark[1]\\
  Dept of Electrical Engineering\\
  Stanford University \\
  \texttt{avoy.datta@stanford.edu} \\
  \And
  Chris Waites \footnotemark[1]\\
  Dept of Computer Science\\
  Stanford University \\
  \texttt{waites@stanford.edu} \\
}

\begin{document}

\maketitle

\begin{abstract}
Data augmentation refers to a wide range of techniques for improving model generalization by augmenting training examples. Oftentimes such methods require domain knowledge about the dataset at hand, spawning a plethora of recent literature surrounding automated techniques for data augmentation. In this work we apply one such method, bilevel optimization, to tackle the problem of graph classification on the \texttt{ogbg-molhiv} dataset. Our best performing augmentation achieved a test ROCAUC score of \textbf{77.77}\% with a GIN+virtual classifier, which makes it the most effective augmenter for this classifier on the leaderboard. This framework combines a GIN layer augmentation generator with a bias transformation and outperforms the same classifier augmented using the state-of-the-art FLAG augmentation.
\end{abstract}

\begin{figure*}[htp!]
    \centering
    \includegraphics[width=\linewidth]{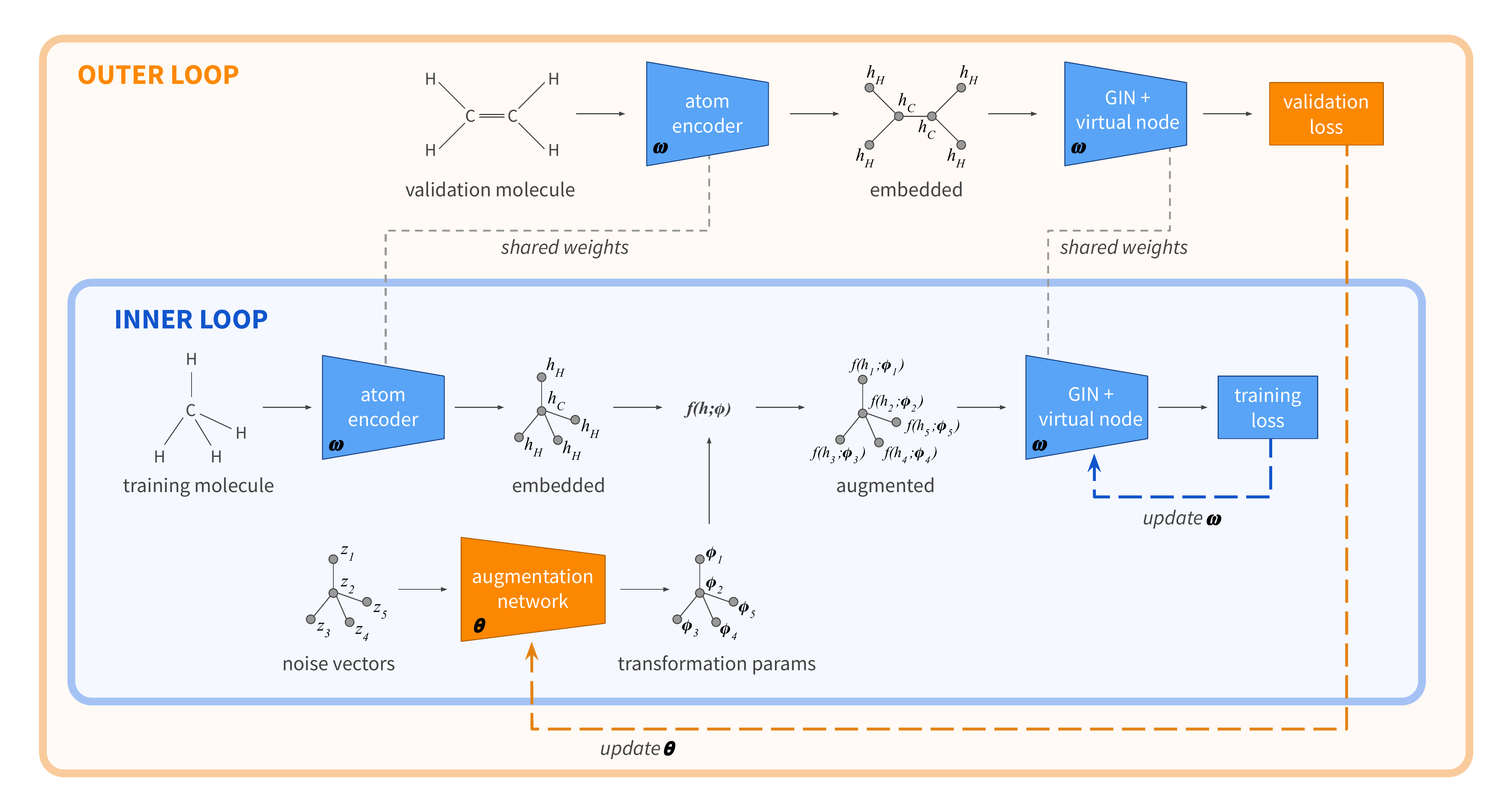}
    \caption{\textbf{GABO Framework.} \textbf{\textit{Inner Loop:}} For every training example, we obtain an atom embedding for each node. We then augment these embeddings using a frozen augmentation generator, and the augmented graph is then passed into the rest of our classifier. Our loss in the inner loop is used to update the parameters $\omega$ of our classifier. Note that we treat the atom encoder and GIN + virtual node as parts of a single classifier, but they are depicted separately here in order to clarify where the augmentation is performed within the forward operation. \textbf{\textit{Outer loop:}} After freezing the graph classifier, we use un-augmented examples from a pseudo-validation set to optimize the parameters $\theta$ of our augmentation generator.}
    \label{fig:architecture}
\end{figure*}
\section{Introduction}
Data augmentation techniques characterize a well known class of approaches used in the context of machine learning to improve model generalization. In this class of techniques, training datasets are expanded by identifying invariances in the training data and exploiting these invariances to generate additional valid training points. 

Automatic data augmentation techniques, broadly speaking, approach this by applying a large set of augmentation maps to a dataset (including those which may not work well), parameterizing them (for example, by the probability of their application), and aiming to find the values for these parameters which optimally improve generalization performance. Although this may produce a good performing data augmentation scheme given a set of known primitive augmentation maps, it still requires a priori knowledge of potential invariances in the data.

The goal of this project is to investigate methods for learning primitive augmentation maps end-to-end from a given classification problem. To the best of our knowledge, this is a new research direction which has not yet seen previous investigation in the context of graphs. Such a data augmentation algorithm could allow for free model generalization improvements for new unknown problems with nuanced invariance structure; this would reduce reliance on expert domain knowledge, proving to be a very important research direction.

In this regard, we have chosen to apply bilevel optimization \cite{mounsaveng2020learning} to the task of automatic data augmentation. Bilevel optimization on a high level describes a metalearning-inspired technique for training which optimizes for the generalization performance of a given model. It performs this by optimizing an augmentation network using the performance of a downstream classification network evaluated on a held out validation set. This is distinct from other  model robustness techniques used in the context of graph classification, for example FLAG \cite{kong2020flag}. FLAG approaches the task of model generalization in the form of adversarial training - that is, it iteratively trains a given model on examples which have small and carefully constructed perturbations which change their label. The assumption is that these small perturbations should not change their label and that the classification model should be robust to such perturbations.

Our work is evaluated on the \texttt{ogbg-molhiv} dataset \cite{hu2021open}, an adaptation from a previous dataset known as MoleculeNet \cite{DBLP:journals/corr/WuRFGGPLP17}. In this setting, each graph represents a molecule in which nodes represent atoms and edges represent bonds between atoms. Node features are 9-dimensional and containing various information pertaining to atoms including their atomic number, chirality, formal charge, and whether or not the atom is in a ring. Molecules are assigned binary labels characterizing whether a molecule inhibits HIV virus replication or not. We believe this to be a relevant problem setting given that manually constructed augmentations are not immediately clear, highlighting the relevance of automatic techniques for data augmentation.

\section{Related Work}

\paragraph{} Our work aims to improve on two existing methods: Free Large-scale Adversarial Augmentation on Graphs (FLAG) and Data Augmentation using Bilevel Optimization (DABO). Our data augmentation could be performed on any graph classification model; we choose to run all of our experiments with GIN virtual, given its position on the leaderboard despite its relatively simple implementation.

\subsection{FLAG}

The FLAG algorithm leverages the concept of ``free'' adversarial training \cite{shafahi2019adversarial_free}. Traditional adversarial training solves the min-max problem: 
\begin{equation}
    \min_\theta \E_{(x, y) \sim D} \left [\max_{||\delta||_p \leq \epsilon} L(f_\theta(x + \delta), y)\right]
\end{equation}

The inner optimization problem is typically solved using Proximal Gradient Descent (PGD), which is strong but inefficent. ``Free" adversarial training cuts down on the runtime by \textit{simultaneously} computing gradients for both the perturbation $\delta$ and the model parameters $\theta$. Optimization over $\delta$ is carried out $M$ times for the same minibatch (ascent step), with the final gradient estimate averaged for robustness. The modified optimization problem is:
\begin{equation}
    \min_\theta \E_{(x, y) \sim D} \left [\frac{1}{M}\sum_{t=1}^M \max_{||\delta||_p \leq \epsilon} L(f_\theta(x + \delta), y)\right]
\end{equation}

To make the optimization tractable in regard to runtime, the authors use \textbf{gradient accumulation} to update the model parameters $\theta$. Essentially, the gradient w.r.t. $\theta$ is computed on every mini-batch, for $M$ times, with the algorithm maintaining a cumulative count of gradients at every step (scaled down by $M$). At the end of each epoch, $\theta$ is updated using the cumulative scaled gradients, obtained by averaging over $M$ iterations of the minibatch.

Some other features FLAG are worth noting:
\begin{itemize}
    \item \textbf{Unbounded attack:} While realistic perturbations for image-based data is usually bounded by some $L-p$ norm on the noise $\delta$, this isn't the case for node features for most graph networks. FLAG addresses this by not imposing an explicit bound on $\delta$ (which is rather implicitly bounded by: $\text{step size }\alpha \times \text{ \# ascending steps }M$)
    
    \item \textbf{Biased perturbation for node classification:} For node inference problems \textit{specifically}, FLAG biases perturbations for far-away nodes using larger step-sizes. The authors justify this approach using superior results in ablation studies.
    
\end{itemize}

\subsection{DABO: Data augmentation with bilevel optimization (computer vision)}


\cite{mounsaveng2020learning} presents a technique for automatic data augmentation map discovery, henceforth referred to as DABO. DABO learns a data augmentation map which, when applied to training data, maximizes the validation accuracy of an end task model. Although, the core limitation is that optimizing a parameterized data augmentation map for this objective is not possible through direct gradient based methods, as the validation loss on unknown holdout data is not a differentiable function of the data augmentation map parameters.

To get around this limitation, they propose an approach which makes use of techniques from online bilevel optimization. Withing this framework, they are able to learn transformations of the training data that minimize the validation loss while training the end task model. On a high level, their training procedure for the augmentation map makes use of an inner loop and an outer loop. In the inner loop, the classifier parameters $\omega$ are trained in the standard supervised way. In the outer loop, the data augmentation parameters $\theta$ are trained on the validation set using an online differentiable method.

\begin{equation}
    \theta \leftarrow \theta - \eta_\theta \nabla_\theta \mathcal{L}(X_{val}, \omega*)
\end{equation}

Given that augmentations by definitions are only applied to the training dataset, it's not immediately clear how the expression $\nabla_\theta \mathcal{L}(X_{val}, \omega*)$ is evaluated. To make it tractable, they make a key observation enabled by the fact that parameters of the classifier $\omega^*$ are shared between the validation and training loss. Given this, we can rewrite this expression as:

\begin{equation}
    \nabla_\theta \mathcal{L}(X_{val}, \omega*) = \frac{\partial \mathcal{L}(X_{val}, \omega^*)}{\partial \omega^*} \frac{\partial \omega^*}{\partial \theta}
\end{equation}

Further, if we define the gradient of the training loss at iteration $t$ as:

\begin{equation}
    \mathcal{G}^{(t)} = \nabla_\omega \mathcal{L}(\mathcal{A}_\theta(X_{train}), \omega^t)
\end{equation}

Then, finally we can write $\frac{\partial \omega^*}{\partial \theta}$ as:

\begin{equation}
    \frac{\partial \omega^*}{\partial \theta} = \sum_{i = 1}^{T - 1} \frac{\partial \omega^{(T)}}{\partial \omega^{(i)}} \frac{\partial \omega^{(i)}}{\partial \mathcal{G}^{(i)}} \frac{\partial \mathcal{G}^{(i)}}{\partial \theta}
\end{equation}

This is sufficient to optimize for using the bilevel optimization procedure described by rolling out the gradients observed during the inner loop. In practice, commonly the last $j$ gradients are used to compute an estimator of the true gradient quantity just described. Additionally, gradient updates are performed on the augmenter every $k$ steps instead of at the end of classifier training, as the latter would be computationally intractable. To the best of our knowledge, this class of augmentations has not been applied in the context of graphs before. Hence, we believe applying this approach to graphs is a novel contribution.

\subsection{GIN with virtual node}

The graph isomorphism network (henceforth referred to as GIN) was first introduced in \cite{DBLP:journals/corr/abs-1810-00826}. This model was first proposed as an investigation of the discriminative power of GNN variants (e.g. graph convolutional network and GraphSAGE). This work was able to show that such models were unable to distinguish simple graph structures. In response to this observation, GIN was developed as a simple architecture that was provably the most expressive among the class of GNNs, being as powerful as the Weisfeiler-Lehman graph isomorphism test. Hence, GIN demonstrates strong empirical performance in the context of many graph classification benchmarks. To briefly characterize this approach, GIN updates node representations as:

\[
    h_v^{(k)} = \text{MLP}^{(k)} \left( \left( 1 + \varepsilon \right) \cdot h_v^{(k - 1)} + \sum_{u \in \mathcal{N}(v)} h_u^{(k-1)} \right)
\]

In the context of graph classification, some approaches will additionally augment graphs by including "virtual" nodes - that is, nodes which are connected to all other nodes in the graph \cite{DBLP:journals/corr/abs-1709-03741, DBLP:journals/corr/GilmerSRVD17, hu2021open}. GIN augmented with virtual nodes acts as the basis for the classification network within our following experimentation.

\section{Methodology}

\subsection{Baseline}

To our knowledge, this is the first application of bilevel optimization to graph neural networks. We have thus decided to compare our model against \textbf{two} prior works: 
\begin{itemize}
    \item \textbf{GIN+virtual node + FLAG:} This model currently has the best performance of all GIN+virtual models on the OGB(\cite{hu2021open}) leaderboard. Given our choice of classifier, and the use of FLAG \cite{kong2020flag} as a state-of-the-art augmentation framework, this provides both a rigorous and fair state-of-the-art we can strive to beat. 
    \item \textbf{Random noise augmentation:} Our second baseline was inspired by the addition of random noise to inputs in computer vision tasks. Experiments have shown that adding small amounts of input noise (jitter) to training data often aids generalization (\cite{noise_addition_jitter_smithing_book}). This has also been mathematically proven to have similar effects on loss objective optimization to Tikhonov Regularization (\cite{tikhonov}). We sample our noise from a uniform distribution capped to the range $[-1, +1]$. 
\end{itemize}

\subsection{GABO: Graph augmentation with bilevel optimization}

Figure \ref{fig:architecture} exhibits our Graph Augmentation framework. Given that bilevel optimization has \textit{not} been previously applied in the context of graphs before, our method is relatively immediate from the work of \cite{mounsaveng2020learning}. We extend their bilevel optimization framework to the task of graph augmentation. Importantly, the augmentation network is tasked with applying a \textit{learned} transformation of the underlying graph to the orinal node features. Obtained this transformation entails consideration of two fundamental design choices: 
\begin{itemize}
    \item \textbf{Inputs to the Augmentation Network:} We use the term \textit{augmentation generation type} to refer to our Augmenter inputs. We considered stochastic inputs (random noise), deterministic graph level features, as well as learned features from a \textit{trainable} GIN Layer (which in turn ingested node features).
    \item \textbf{Functional form of Augmentation}: We refer to this using \textit{node embedding transformation type}. This defines the degree to which the augmentation is allowed to impact the original node features. We used a simple bias addition, an elementwise multiplication transform and a \textit{shifted} elementwise multiplication transform for experiments.
\end{itemize}

\subsection{Augmentation generation types}

Given that the Augmenter is responsible for sampling augmentation parameters, it remains a question what source of randomness is fed in and what information about the current batch the Augmenter is aware of. In this respect, throughout our experimentation we used three augmentation generation types, including: simple random noise, classic node features, and GIN features. Simple random noise corresponds to feeding the Augmenter purely random inputs, in particular randomly sampled noise from the uniform distribution in the range $[-1, 1]$. Classic node features corresponds to sampling noise as before but also concatenating deterministic graph level features to the noise vector, including measures of the betweenness centrality, closeness centrality, degree, and pagerank for each node. Finally, we also include a setting in which learned features from a trainable GIN Layer are concatenated to the noise vector and fed into the Augmenter, incorporating node-level feature information.

\subsection{Node embedding transformation types}

The augmentation generator outputs the parameters of a transformation, which is performed on nodes of training graphs. These transformations are written as $f(h;\phi)$ in Fig.\ref{fig:architecture}, where $h \in \mathbb{R}^d$ is the atom embedding of a node. We implement the following three types of transformations: 
\begin{itemize}
    \item Bias: $f(h;\phi) = h + \phi$
    \item Element-wise multiplication (with bias): $f(h;\phi) = \phi_1\odot h + \phi_2$
    \item Shifted element-wise multiplication (with bias): $f(h;\phi) = (1+\phi_1)\odot h + \phi_2$
\end{itemize}
Note that we choose to perform element-wise multiplication rather than an affine transformation, because the latter would require the augmenter to generate a $d\times d$ weight matrix $\phi_1$. This would be computationally prohibitive given our resources for this project, and we believe we'd risk overfit on our pseudo-validation set.


\section{Experimental Setup}

\subsection{Datasets}

GABO requires partitioning the original full dataset into four splits (train, pseudo-validation, validation, and test) instead of the traditional three. We compute validation loss on the pseudo-validation set during the outer loop of our bilevel optimization, while our validation set is used to perform early stopping and model selection.

Throughout our experimentation, we try a number of distinct dataset splitting schemes to perform this partition. These are two fold - scaffolding, and randomized. The scaffold splitting attempts to separate molecules which differ structurally into distinct subsets, providing a more realistic estimate of model performance in experimental settings. The randomized method simply shuffles molecules plainly among the train, validation, and pseudo-validation sets, leaving the test set consistent with the OGB leaderboard. Although in theory the scaffolding approach would be expected to yield a more difficult learning setting, in practice we find that distinctions in dataset splitting do not yield significant differences in outcome.

Since GABO requires four data splits instead of three, our validation set is smaller than it would be using a traditional split, inducing more variance in our results. We, therefore, select three models which perform best on our validation set and report test performance for all of them in our performance table.

\subsection{An important note about optimization}

Though the models on the leaderboard which serve as our baselines are trained using the Adam optimizer (\cite{kingma2017Adam}), the use of the \texttt{pytorch meta} package restricted us to using a Stochastic Gradient Descent (SGD) with momentum for updates to both Augmenter and Classifier networks. \texttt{pytorch meta} is fairly new and so far does not support functionality for Adam. While SGD can sometimes generalize better than Adam for vision tasks (\cite{zhou2020theoretically_SGD_better}), Adam has been shown to perform consistently better than SGD on loss landscapes with sharp local minima (\cite{Adam_better_wilson2018marginal}). We therefore have reason to believe we could have gotten higher scores had we had access to an Adam optimizer implementation within \texttt{pytorch meta}, and we leave this open as an avenue for future exploration. 

\subsection{Scheduling and hyperparameters}
\label{subsec: sch_and_hparams}
Each model is trained over a total period of 200 epochs, with a patience hyperparameter of 30 epochs for early stopping. The initial learning rate is set to 0.1, and we use a stepwise learning rate scheduler set to scale the initial rate with a constant factor $\frac{1}{5}$ over the schedule $[60, 120, 160]$. We also use L2 regularization over \textit{both} augmenter and classifier loss objectives, with regularization constants of 0.01 for the augmenter and 5e-4 for the classifier. Additionally, we use a latent dimension for the random noise input into our augmenter of 10. Our node embeddings have dimension 256.

\subsection{Computational Power}

We ran a total of \textbf{53} experiments with the GABO framework, cumulating a total of roughly \textbf{165 GPU hours}. All experiments were run on Tesla K80 GPUs on Amazon Web Services (AWS) \texttt{p2.xlarge} instances.

\section{Results and Discussion}

Our main result is that the best-performing GABO model, which combines a GIN embedding augmentation generator with a bias embedding transformation, outperforms GIN with virtual node and FLAG augmentation. This places us above all other GIN entries in the leaderboard.

\begin{table}[htb]
    \centering
    \begin{tabular}{c c c | c c c} 
    \hline \hline 
    Method & Transform & Optimizer & Validation & Test \\
    \hline 
    GIN+Virtual & - & Adam & 84.79 $\pm$ 0.68 & 77.07$\pm$1.5  \\
    GIN+Virtual+FLAG & Bias & Adam & 84.38 $\pm$ 1.3 & 77.48 $\pm$ 0.96 \\
    \hline
    Baseline & Bias & SGD+mom & 68.70$\pm$0.26 & 68.88$\pm$0.42 \\
    \hline
    GABO w/ \textbf{Noise Input} & Bias & SGD+mom & 84.83$\pm$1.2 & 76.17$\pm$2.0 \\
    GABO w/ \textbf{Classic Features} & Bias & SGD+mom & 82.24$\pm$1.2 & 77.43$\pm$0.82 \\
    GABO w/ \textbf{GIN Embeddings} & Bias & SGD+mom & 81.15$\pm$3.3 & \textbf{77.77$\pm$.40} \\
    \hline \hline \\
    \end{tabular}
    \caption{\textbf{Performance Table}}
    \label{tab:perf}
\end{table}

\begin{figure}[htb]
    \centering
    \includegraphics[width=\textwidth]{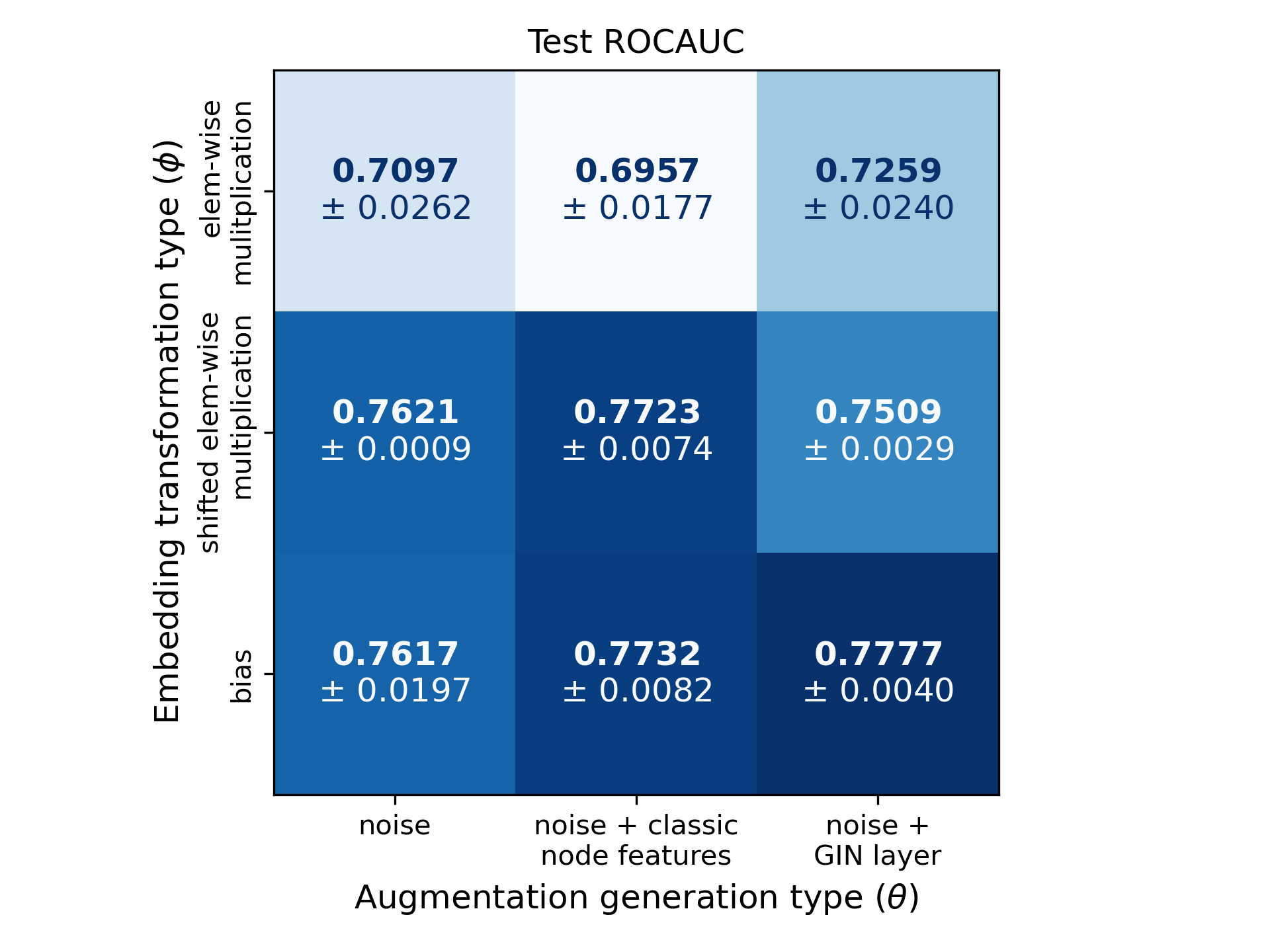}
    \caption{Ablation study on the effects on transformation type and augmentation generation type on test performance. Given our limited computing capabilities, we perform training with five seeds for only experiments in the \textbf{bottom row} of this matrix. For all other cells, mean and standard deviation for ROCAUC are computed over experiments with different patience hyperparameters, which we determined did not make a significant difference on validation performance. We recognize that this is not a traditional method for computing mean and standard deviation of performance metrics, and, given more computing resources, we would have opted to run five seeds on all cells.}
    \label{fig:cmap}
\end{figure}

In the following sections, we describe trends observed in the rest of our experiments. Performance for different combinations of our augmentation generation types with our transformation types is summarized in Fig.~\ref{fig:cmap}.

\subsection{Baseline performance}

Our baseline, augmenting with random noise bias, performed poorly compared to leaderboard entries. It is clear that this model underfit, since performance on the training set similarly lags behind models with an augmentation network. The authors in \cite{tikhonov} showed that the standard deviation of noise added is equivalent to the regularizing factor in Tikhonov regularization; we suspect this underfitting is due to over-regularization at choosing too large a standard deviation for the baseline noise distribution. This would lead to the magnitude of noise added to node embeddings hampering training. This over-regularization in our baseline highlights one of the benefits of our approach, which directly optimizes the bias to perform well on unseen data.

\subsection{Effect of augmentation generation type}

Of the three augmentation generation types we implemented, GIN layer with noise performs best, followed by classic node features with noise, then noise on its own. This makes intuitive sense, since the GIN layer augmenter makes use of node input features while classic node features do not. 

One thing to note is that for all experiments where GIN is employed as an augmentation generator, we use a \textit{single} GIN layer. This means transformation parameters are generated only with information from a node's immediate neighbors (1-hop neighborhoods). Future work on this project could include exploring deeper augmentation generators, which would capture information from larger k-hop neighborhoods.

\subsection{Effect of transformation types}

Of the three transformation functional forms, adding the augmentations as a bias term gives us the best performance. In general, performance steadily peaks for the transformation type set to ``\texttt{bias}" for all ``\texttt{augmentation generation}" types (figure \ref{fig:cmap}, reading top to bottom).  This makes slightly less intuitive sense than the preceding subsection, as we \textit{expected} the model to learn effective dimension-specific scale factors for the original input node embeddings. In \textit{reality}, our experiments tell us the optimal way to apply augmentations to input node embeddings with GABO is to treat them as a bias term. 

\begin{figure}[htb!]
    \centering
    \includegraphics[width=\textwidth]{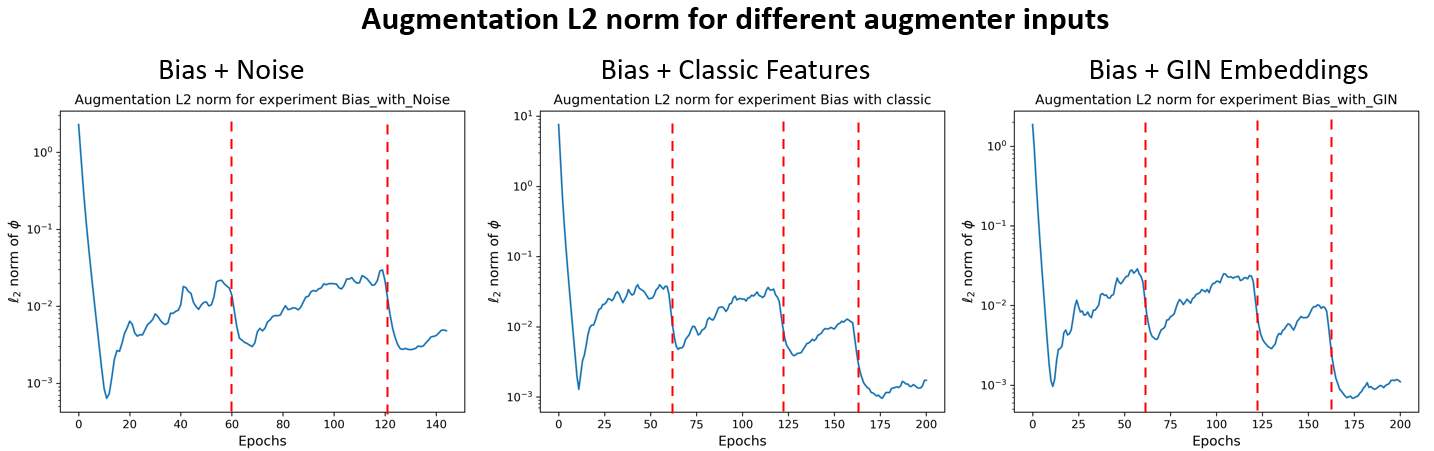}
    \caption{L2-norm of the augmentation vector over time. The red vertical lines denote epochs when the learning rate is dropped.}
    \label{fig:phi_training}
\end{figure}

\subsection{Analysis of transformation weights ($\phi$) over time}

Unlike augmentations in computer vision, graph augmentations are harder to visualize and provide intuition for at inference time. We can, however, gauge how significant the augmentations are by looking at the L2 norm of the $\phi$ vector over time (figure \ref{fig:phi_training}). Note that our model learns non-trivial bias vectors, as they increase in magnitude with more training. An interesting trend we noted is the L2 norms seem to \textbf{drop} every time the learning rate scheduler scales the learning rate (by $\frac{1}{5}$, as described in subsection \ref{subsec: sch_and_hparams}). 

\section{Conclusion}

In this work we show that bilevel optimization can be used to improve upon graph classification results for a relevant baseline on the \texttt{ogbg-molhiv} dataset. Further, this augmentation procedure can be followed with no a priori domain knowledge about the task at hand. Throughout our investigation we evaluate the efficacy of various design choices in our approach, including choices in augmentation generation types, transformation types, and data splitting schemes. We find that the best performing GABO framework combines a GIN layer augmentation generator with a bias transformation which outperforms GIN with virtual node and FLAG augmentation.

Going forward there exist a number of interesting directions for future work. As referenced previously, the augmentation network is currently optimized using SGD due to the constraints of the metalearning framework (\texttt{pytorch-meta}) we use. Given the strong empirical performance of the Adam optimizer, we believe swapping SGD for Adam could yield a reasonable performance improvement. In addition, in our experiments we only use GIN with a single layer. Although we provide graph-level information in the form of classic node features for certain experiments, increasing the depth of GIN to consider multi-hop information might improve performance. Furthermore, experimenting with other augmentation types beyond linear operations and other classification tasks for our dataset and other datasets would be worthwhile in future investigation.




\bibliography{example_paper}
\bibliographystyle{icml2020}

\end{document}